\documentclass[runningheads]{llncs}
\usepackage[T1]{fontenc}
\usepackage{color}
\usepackage{graphicx,verbatim}
\usepackage{amsfonts}
\usepackage{amssymb}
\usepackage{amsmath}
\usepackage{booktabs}
\usepackage{adjustbox}
\usepackage[colorlinks=true,linkcolor=red,citecolor=blue,urlcolor=blue]{hyperref}
\usepackage{pifont}
\usepackage[table]{xcolor}
\usepackage{graphicx}
\usepackage{marvosym}

\begin{document}
\title{VAMPIRE: Uncovering Vessel Directional and Morphological Information from OCTA Images for Cardiovascular Disease Risk Factor Prediction}
\titlerunning{VAMPIRE}

\author{Lehan Wang\inst{1} \and
Hualiang Wang\inst{1}\and
Chubin Ou\inst{2}$^($\textsuperscript{\Letter}$^)$ \and
Lushi Chen\inst{3} \and
Yunyi Liang\inst{4} \and
Xiaomeng Li\inst{1}$^($\textsuperscript{\Letter}$^)$
}

\def\etal{\emph{et al.}}
\def\vs{\emph{v.s. }}

\authorrunning{L. Wang et al.}

\institute{The Hong Kong University of Science and Technology, Hong Kong, China \\
\email{eexmli@ust.hk} \\
\and
Department of Radiology, Guangdong Provincial People's Hospital (Guangdong Academy of Medical Sciences), Southern Medical University, Guangzhou, China \\
\email{cou@connect.ust.hk} \\
\and
Health Management Center, Foshan First People's Hospital
\and
Health Management Center, The Sixth Affiliated Hospital, School of Medicine, South China University of Technology
}

\newcommand{\xmli}[1]{{\color[rgb]{1,0.0,0.1}{[XM: #1]}}}
\newcommand{\lh}[1]{{\color[rgb]{0.5,0.3,0.8}{[LH: #1]}}}
\newcommand{\yq}[1]{{\color[rgb]{0.1,0.1,0.8}{[YQ:#1]}}}
    
\maketitle              
\begin{abstract}
Cardiovascular disease (CVD) remains the leading cause of death worldwide, requiring urgent development of effective risk assessment methods for timely intervention. 
While current research has introduced non-invasive and efficient approaches to predict CVD risk from retinal imaging with deep learning models, the commonly used fundus photographs and Optical Coherence Tomography (OCT) fail to capture detailed vascular features critical for CVD assessment compared with OCT angiography (OCTA) images. 
Moreover, existing methods typically classify CVD risk only as high or low, without providing a deeper analysis on CVD-related blood factor conditions, thus limiting prediction accuracy and clinical utility. 
As a result, we propose a novel multi-purpose paradigm of CVD risk assessment that jointly performs CVD risk and CVD-related condition prediction, aligning with clinical experiences. 
Based on this core idea, we introduce OCTA-CVD, the first OCTA dataset for CVD risk assessment, and a \textbf{V}essel-\textbf{A}ware
\textbf{M}amba-based \textbf{P}rediction model with \textbf{I}nfo\textbf{r}mative \textbf{E}nhancement (\textbf{VAMPIRE}) based on OCTA enface images. Our proposed model aims to extract crucial vascular characteristics through two key components: (1) a Mamba-Based Directional (MBD) Module that captures fine-grained vascular trajectory features and (2) an Information-Enhanced Morphological (IEM) Module that incorporates comprehensive vessel morphology knowledge. Experimental results demonstrate that our method can surpass standard classification backbones, OCTA-based detection methods, and ophthalmologic foundation models. Our codes and the collected OCTA-CVD dataset are available at \href{https://github.com/xmed-lab/VAMPIRE}{https://github.com/xmed-lab/VAMPIRE}.


\end{abstract}

\section{Introduction}

\begin{figure*}[t]
	\centering
	\includegraphics[width=1.0\textwidth]{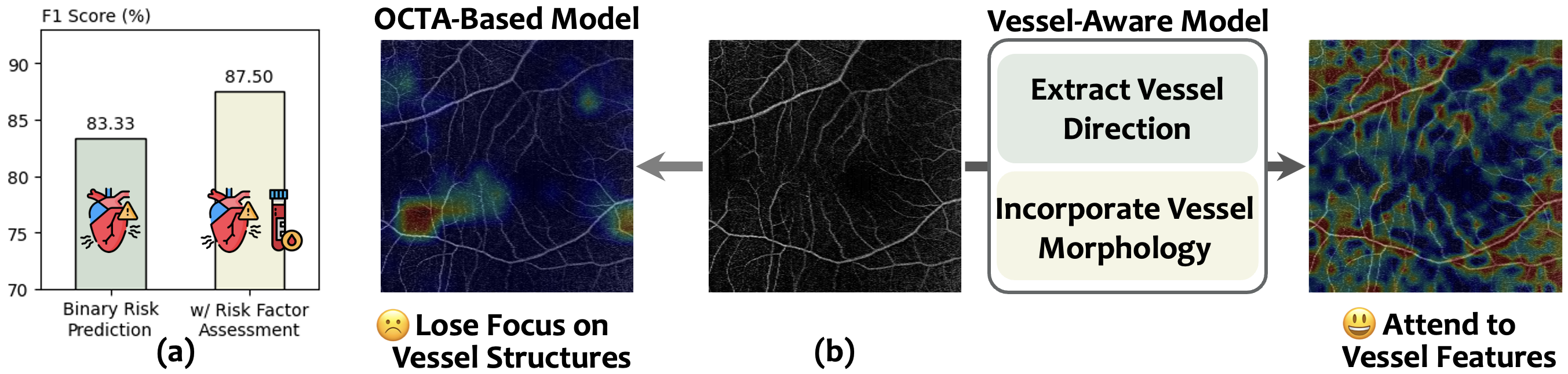}
	\caption{Comparative Analysis of (a) Task Paradigm and (b) Model Capability.
    (a) Performance Comparison between binary risk classification and joint risk factor prediction. (b) Evaluation of vessel feature representation abilities between OCTA-Based Model and our proposed Vessel-
Aware Model.}
	\label{fig:intro}
\end{figure*}

Cardiovascular diseases (CVDs) remain a significant burden on public health and a leading cause of mortality worldwide. The World Health Organization (WHO) estimates that CVDs are responsible for nearly 18 million deaths each year, accounting for over 30\% of all global deaths~\cite{cesare2024heart}. Thus, risk assessment of CVD is critical, as early intervention can prevent severe cardiovascular events.

Recent advances in retinal imaging and artificial intelligence (AI)~\cite{chen2025mutri,li2024vessel,li2020self,li2021rotation} have the potential to revolutionize cardiovascular risk assessment into a non-invasive and automatic approach. The retinal vascular structure shares anatomical similarities with the coronary circulation, making it a promising biomarker for monitoring vascular health~\cite{farrah2019retinal,seidelmann2016retinal}. Several studies~\cite{cheung2021deep,diaz2022predicting,lee2023multimodal,maldonado2024predicting,poplin2018prediction} have leveraged deep learning models for fundus photography or Optical Coherence Tomography (OCT) to predict CVD risk based on retinal vascular morphology and structural changes.
Poplin \etal~\cite{poplin2018prediction} first attempted to apply a deep learning model to fundus photos for predicting the risk of CVD events. Cheung \etal~\cite{cheung2021deep} automatically measured retinal-vessel calibre in retinal photographs and demonstrated its association with CVD risk factors.
Other works~\cite{diaz2022predicting,lee2023multimodal} further complemented fundus photos with additional modalities, like cardiovascular magnetic resonance images and clinical risk factors, to further improve the risk prediction accuracy. Maldonado \etal~\cite{maldonado2024predicting} developed an AI model with OCT scans to leverage subtle abnormalities in retinal microstructure.
These methods have achieved promising result in non-invasive CVD risk prediction, serving as a valuable alternative to conventional CVD risk assessment approaches and facilitating earlier detection. 

Despite these advancements, two significant limitations hinder these models in providing more reliable and accurate risk prediction.
First, the input modalities employed in the previous methods, fundus photography and OCT, capture insufficient vascular details for precise CVD risk assessment due to their limited resolution of microvascular structures, particularly compared with OCT angiography (OCTA) images that reflect capillary networks and vessel density more clearly. 
Second, existing approaches merely perform binary CVD risk detection without assessing fine-grained CVD risk factors, such as blood glucose and cholesterol, which undermines both confidence and reliability.
As proved by~\cite{wong2006retinal}, these blood biomarkers are vital in determining CVD risk, and excluding these factors would notably reduce risk prediction accuracy. Additionally, simply providing binary risk classification (high/low risk) poses challenges for patients to understand and trust the results, preventing the implementation of timely and personalized intervention strategies.

To overcome these limitations, we propose a novel paradigm of utilizing OCTA images to predict four key CVD-related conditions, including high blood glucose, high blood cholesterol, high blood triglycerides, and high blood pressure, while jointly performing 10-year CVD risk prediction. 
Unlike existing methods that solely perform binary risk classification, our multi-task framework provides deeper insights into blood biomarker abnormalities, which align well with empirical clinical knowledge, thus substantially enhancing CVD risk prediction performance by over 4\% in F1 score, see in Fig.~\ref{fig:intro} (a). 
Besides, the identification of blood factor conditions offers more clinically interpretable results could be delivered, thereby addressing a critical gap in current CAD risk assessment frameworks.

To fulfill this objective, we collect \textbf{the first OCTA enface dataset, OCTA-CVD, for joint CVD risk and factor estimation}.
{We first evaluate existing deep learning models developed to extract modality-specific features from OCTA images.~\cite{hao2024early,liu2024beyond,wang2022screening,zhao2023eye}.}
However, our evaluation reveals that directly applying these methods inadequately captures continuous vascular structures, which is a crucial characteristic in OCTA imaging, as illustrated in Fig.~\ref{fig:intro} (b).
Consequently, we draw inspiration from Mamba's capability in modeling long-range dependencies~\cite{gu2023mamba,hatamizadeh2024mambavision,wang2024serp,vim}, which is suitable for capturing vascular trajectory. 
Nevertheless, Mamba-based models typically process images in patches, hindering the comprehensive understanding of continuous morphological features.
Building upon these insights, we introduce a \textbf{V}essel-\textbf{A}ware \textbf{M}amba-based \textbf{P}rediction model with \textbf{I}nfo\textbf{r}mative \textbf{E}nhancement (\textbf{VAMPIRE})
for joint CVD risk and CVD-related factor estimation given OCTA enface images. The proposed VAMPIRE model consists of two components: \ding{182} \textbf{Mamba-Based Directional (MBD) Module} focusing on the intricate trajectories of inner vascular paths, and \ding{183} \textbf{Information-Enhanced Morphological (IEM) Module} to integrate comprehensive vessel shape information. 
These two components could collaboratively extract CVD-specific vessel features from OCTA enface images, thus effectively leveraging vascular structure information for accurate prediction of CVD risk and related factors. By combining both directional and morphological knowledge, our method could effectively attend to vessel features as in Fig.~\ref{fig:intro} (b), and outperform general classification baselines, OCTA-based detection models, and ophthalmological foundation models on multi-center validation datasets.
\section{Methodology}

\begin{figure*}[t]
	\centering
	\includegraphics[width=0.75\textwidth]{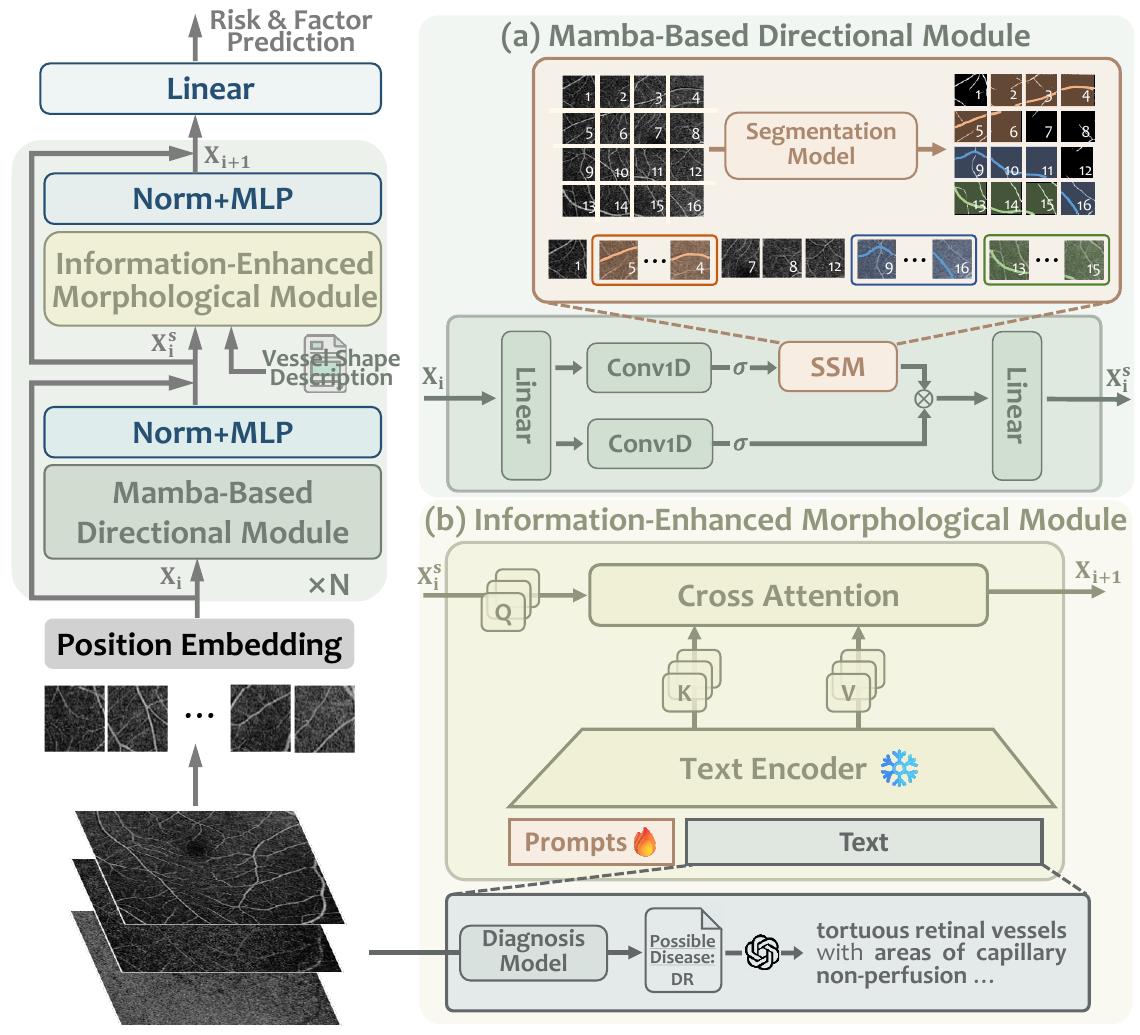}
	\caption{Overview of the Proposed VAMPIRE Framework, which contains (a) Mamba-Based Directional (MBD) Module for vascular trajectory feature extraction and (b) Information-Enhanced Morphological Module for vessel shape knowledge integration.}
	\label{fig:overview}
\end{figure*}

\subsection{Architecture Overview}
An overview of our framework is presented in Fig.~\ref{fig:overview}. To align with standard Mamba sequence processing, we first segment the input OCTA enface image $\mathbf{I} \in \mathbb{R}^{B \times H \times{W}}$ into flattened patches $\mathbf{P} \in \mathbb{R}^{B \times N \times h \times w}$, where $B$ represents the batch size, $N$  denotes the number of patches, and $(H, W)/(h,w)$ corresponds to the size of original image/patch. These 2D patches are then projected into 1D sequences and combined with position encodings, yielding $\mathbf{X} \in \mathbb{R}^{B \times N \times D}$, where $D$ represents the dimension of patch embeddings. Following Vim~\cite{vim}, we also apply the class token $\mathbf{t}_{cls}$ to encode global knowledge. Next, the complete patch sequence $\mathbf{X_0}=[\mathbf{t}_{cls}; \mathbf{X}]$ is fed into the proposed sequential VAMPIRE blocks, which is composed of two parts: a Mamba-Based Directional (MBD) Module that effectively extracts vascular sequence features (\S~\ref{sec:mbd}), and an Information-Enhanced Morphological (IEM) Module that incorporates vessel morphology descriptions to enrich the model's representation (\S~\ref{sec:iem}). Finally, the extracted features are input into a linear classification layer to obtain the final prediction for CVD risk and related conditions. The model is optimized with binary cross-entropy loss as the classification loss function. In the following sections, we will elaborate on the details of each constituent module.

\subsection{Mamba-Based Directional Module}
\label{sec:mbd}
To effectively extract vessel trajectory features from OCTA enface images, we adopt the Mamba-Based Directional (MBD) Module for modeling vascular sequences, as demonstrated in Fig.~\ref{fig:overview} (a). 
The intricate structure of blood vessels in OCTA images poses challenges for conventional Mamba scanning operations. The commonly applied linear- or cross-scanning approaches inevitably disrupt vessel continuity, thereby constraining the model's capability to capture intrinsic vascular structure.
To address this limitation, we intuitively introduce vessel segmentation maps as directional guidance to formulate the scanning order within the State Space Model (SSM) module.
Specifically, we employ SAM-OCTA~\cite{wang2024sam}, a pre-trained vessel segmentation network for OCTA images, to generate initial vessel maps, followed by noise filtering and morphological transformation to acquire well-defined vascular structures. Subsequently, we use depth-first search (DFS) to traverse the vessel tree in the segmentation map to obtain different continuous trajectories of large vessels, and generate patch sequences along each vascular path $\mathbf{V_i}=[v_{i,1},...v_{i,n_i}]$ based on the refined segmentation map. 
The complete scanning sequence is then constructed by concatenating vessel patches from different branches with intermediate background patches $\mathbf{B_{i,j}}=[b_{i,1},...b_{i,m_i}]$, which can be formulated as:
\begin{equation}
    \mathbf{S} = \text{concat}_{i,j \in T} [\mathbf{V_i};\mathbf{B_{i,j}};\mathbf{V_j}] \:,
\end{equation}
where $T$ is the number of all the continuous vessel trajectories.

Through the vessel-following traversal paths, our proposed MBD module enables each OCTA patch to integrate contextual information from adjacent patches along the same vascular branch, thus encouraging the model to concentrate on CVD relevant vascular features. 

\subsection{Information-Enhanced Morphological Module}
\label{sec:iem}
After obtaining the output of the MBD module, $\mathbf{X_i^s}$, we proceed to integrate vessel shape descriptive information through our subsequent Information-Enhanced Morphological (IEM) module. 
Considering that the patch-wise encoding mechanism in Mamba-based architecture would potentially constrain the model's ability to comprehend global vascular morphology, we propose to promote feature extraction through textual descriptions generated by the Multimodal Large Language Model (MLLM), as shown in Fig.~\ref{fig:overview} (b).
However, we observe that directly feeding OCTA images into MLLMs yields generic and uninformative descriptions, such as ``the retinal vessels are well-defined'', providing limited morphological insights.
Consequently, we first employ a classification model trained on the OCTA-500 dataset~\cite{li2024octa} to identify potential retinal diseases. Subsequently, we prompt GPT-4o~\cite{hurst2024gpt} with the diagnostic results to generate descriptions $\mathbf{t}$ on possible vascular morphologies. Combined with learnable prompts $\mathbf{p}$, these descriptions are fed into a frozen text encoder $\Theta_T$ and then integrated with image representation $\mathbf{X_i^s}$ through the cross-attention mechanism. The process is denoted as:
\begin{align}
    \mathbf{T} &= \Theta_T([\mathbf{p};\mathbf{t}]) \:, \\
    \mathbf{X_{i+1}}&=\text{CrossAttention}(\mathbf{X_i^s},\mathbf{T},\mathbf{T}) \:,
\end{align}

Through this approach, our model could effectively capture both fine-grained vessel trajectory features and global morphological information, enhancing its capability to identify CVD-associated microvascular patterns.

\section{Experiments}
\subsection{Experimental Setup}
\textbf{Dataset.} 
We collect an in-house OCTA dataset, OCTA-CVD\footnote[1]{Our OCTA-CVD dataset is released at \href{https://github.com/xmed-lab/VAMPIRE}{https://github.com/xmed-lab/VAMPIRE}}, for joint CVD risk and factor estimation, which contains 1659 OCTA enface images of Superficial (ILM-IPL), Deep (INL-OPL), and Avascular (OPL-BM) layers from 843 patients in a local hospital’s health check-up center from 2022 August to 2022 December. {The dataset includes 74.3\% female and 25.7\% male, with a mean age of 45$\pm$12 years.}. The OCTA images were captured in a 6$\times$6 mm$^2$ area centering at fovea with a Velite C3000 OCTA device. Each patient is annotated with CVD High Risk (HR) and CVD-related conditions derived from blood test, including high blood glucose (HG), high blood cholesterol (HC), high blood triglycerides (HTG), and high blood pressure (hypertension, HTN).
The data statistics is shown in Table~\ref{tab:data-statistics}. The dataset is divided using five-fold cross-validation, maintaining consistent class distributions in both training and test sets. To prevent data leakage, we ensured that data from individual patients appeared in only one subset.

\begin{table}[t]
    \caption{Data Statistics of the collected OCTA-CVD dataset.}
    \label{tab:data-statistics}
    \centering
    \begin{adjustbox}{width=0.45\textwidth}
        \begin{tabular}{ccccccc}
        \toprule
        \textbf{Category} & \textbf{HR} & \textbf{HG} & \textbf{HC} & \textbf{HTG} & \textbf{HTN}& \textbf{Total} \\ 
        \midrule
        \textbf{Patients} & 101 & 81 & 267 & 296 & 215 & 843 \\
        \textbf{Images} & 195 & 160 & 525 & 580 & 420 & 1659 \\
        \bottomrule
        \end{tabular}
    \end{adjustbox}
\end{table}

\noindent\textbf{Data Preprocessing.} We first apply median filter for OCTA enface images for noise reduction, and then implement contrast-limited adaptive histogram equalization (CLAHE) to enhance vessel clarity. All the
images are resized to 448$\times$448 for input. We adopt data augmentation including random crop, flip, and rotation.

\noindent\textbf{Implementation Details.} 
During training, we also incorporate patient demographic information (age and gender) as additional input features to concatenate with the image features for final classification. The OCT layers are concatenated along the channel dimension to form a three-channel input.
All the experiments are conducted on the NVIDIA RTX 3090 GPU. We employ AdamW optimizer with a base learning rate of $1e^{-4}$, scaled according to batch size, and a weight decay of $5e^{-2}$. A cosine annealing scheduler is applied to gradually reduce the learning rate to zero. For fair evaluation, we apply identical data preprocessing strategies and running epochs for all the experiments and report the average performance of all the categories at the final epoch across five-fold cross-validation. Our results are reported at eye level.

\subsection{Comparison with State-of-the-Art Models}
To demonstrate the effectiveness of our proposed approach, we compare our model with Convolutional Neural Network (CNN) backbones~\cite{dosovitskiy2020image,he2016deep,liu2022swin,tan2019efficientnet,vim}, OCTA-based detection methods~\cite{hao2024early,liu2023polar}, and fine-tuned ophthalmologic foundation models~\cite{qiu2024development,zhou2023foundation}. 
As shown in Table~\ref{tab:main-results}, it is apparent that fine-tuned ophthalmologic foundation models can outperform both CNN backbones and OCTA-based models. This suggests that the inherent ophthalmologic knowledge facilitates the model to effectively adapt to the OCTA data distribution even without specific pre-training on OCTA images. 
Furthermore, the comparison results demonstrate that our proposed VAMPIRE model marginally surpasses all the other baselines. Notably, our approach achieves the improvement of 7.55\% (64.67\% \vs 57.12\%) and 11.6\% (64.67\% \vs 53.07\%) compared with the best AUPR of CNN backbones and OCTA-specific models. Additionally, our model improves the state-of-the-art fine-tuned ophthalmologic foundation model from 55.22\% to 62.71\% in F1 score and 60.09\% to 64.67\% in AUPR. 

\begin{table*}[t]
    \caption{Results on our collected OCTA-CVD dataset. The bold and underlined numbers indicate the best and second-best score, respectively. $\dagger$: CNN backbones; $\star$: OCTA-based detection methods; $\ddagger$: Fine-tuned ophthalmologic foundation models.}
    \label{tab:main-results}
    \centering
    \begin{adjustbox}{width=1.0\textwidth}
        \begin{tabular}{lcccccc}
        \toprule
        \textbf{Model} & \textbf{Precision} & \textbf{Recall} & \textbf{F1 Score} & \textbf{Accuracy} & \textbf{AUC} & \textbf{AUPR} \\
        \midrule
        ResNet$^\dagger$\cite{he2016deep} & 0.4928\scriptsize{$\pm$0.0396} & 0.5750\scriptsize{$\pm$0.0387} & 0.5225\scriptsize{$\pm$0.0268} & 0.7576\scriptsize{$\pm$0.0244} & 0.7581\scriptsize{$\pm$0.0244} & 0.5295\scriptsize{$\pm$0.0224}  \\
        
        EfficientNet$^\dagger$\cite{tan2019efficientnet} & 0.4581\scriptsize{$\pm$0.0387} & 0.6057\scriptsize{$\pm$0.0542} & 0.5188\scriptsize{$\pm$0.0352} & 0.7485\scriptsize{$\pm$0.0179} & 0.7582\scriptsize{$\pm$0.0215} & 0.5207\scriptsize{$\pm$0.0340}  \\
        
        ViT$^\dagger$\cite{dosovitskiy2020image} & 0.5058\scriptsize{$\pm$0.0593} & 0.5783\scriptsize{$\pm$0.0270} & 0.5374\scriptsize{$\pm$0.0214} & 0.7752\scriptsize{$\pm$0.0202} & 0.7778\scriptsize{$\pm$0.0222} & 0.5513\scriptsize{$\pm$0.0164}  \\
        
        Swin V2$^\dagger$\cite{liu2022swin} & 0.5010\scriptsize{$\pm$0.0279} & 0.5592\scriptsize{$\pm$0.0589} & 0.5265\scriptsize{$\pm$0.0264} & 0.7782\scriptsize{$\pm$0.0282} & 0.7879\scriptsize{$\pm$0.0356} & 0.5582\scriptsize{$\pm$0.0357}  \\
        
        Vim$^\dagger$\cite{vim} & 0.4636\scriptsize{$\pm$0.0137} & 0.5517\scriptsize{$\pm$0.0461} & 0.4851\scriptsize{$\pm$0.0161} & 0.7818\scriptsize{$\pm$0.0450} & 0.7971\scriptsize{$\pm$0.0263} & 0.5712\scriptsize{$\pm$0.0308} \\
        
        PolarNet$^\star$\cite{liu2023polar} & \underline{0.5758\scriptsize{$\pm$0.0376}} & 0.4142\scriptsize{$\pm$0.0341} & 0.4077\scriptsize{$\pm$0.0509} & 0.7788\scriptsize{$\pm$0.0043} & 0.7569\scriptsize{$\pm$0.0474} & 0.5291\scriptsize{$\pm$0.0124} \\
        
        Eye-AD$^\star$\cite{hao2024early} & 0.5153\scriptsize{$\pm$0.0199} & 0.6349\scriptsize{$\pm$0.0132} & \underline{0.5754\scriptsize{$\pm$0.0102}} & 0.7594\scriptsize{$\pm$0.0126} & 0.7501\scriptsize{$\pm$0.0235} & 0.5307\scriptsize{$\pm$0.0143} \\ 
        
        RETFound$^\ddagger$\cite{zhou2023foundation} & {0.5614\scriptsize{$\pm$0.0077}} & 0.5886\scriptsize{$\pm$0.0356} & 0.5407\scriptsize{$\pm$0.0116} & \underline{0.7885\scriptsize{$\pm$0.0116}} & 0.8024\scriptsize{$\pm$0.0190} & 0.5838\scriptsize{$\pm$0.0382} \\
        
        VisionFM$^\ddagger$\cite{qiu2024development} & 0.5610\scriptsize{$\pm$0.0355} & \underline{0.6351\scriptsize{$\pm$0.0439}} & 0.5522\scriptsize{$\pm$0.0255} & 0.7739\scriptsize{$\pm$0.0164} & \underline{0.8061\scriptsize{$\pm$0.0192}} & \underline{0.6009\scriptsize{$\pm$0.0341}} \\
        
        \textbf{VAMPIRE} & \textbf{0.6068\scriptsize{$\pm$0.0652}} & \textbf{0.6570\scriptsize{$\pm$0.0442}} & \textbf{0.6271\scriptsize{$\pm$0.0293}} & \textbf{0.8012\scriptsize{$\pm$0.0183}} & \textbf{0.8244\scriptsize{$\pm$0.0153}} & \textbf{0.6467\scriptsize{$\pm$0.0394}} \\
        
        \bottomrule
        \end{tabular}
    \end{adjustbox}
\end{table*}

\begin{table*}[t]
    \centering
    \begin{minipage}[t]{0.49\textwidth}
    \centering
    \caption{Extended Comparison on Supplementary Dataset. The AUC metric for each category is presented.}
    \begin{adjustbox}{width=\linewidth}
        \begin{tabular}{lcccccc}
        \toprule
        \textbf{Model} & \textbf{HR} & \textbf{HG} & \textbf{HC} & \textbf{NTG} & \textbf{HTN} \\
        \midrule
        ViT~\cite{dosovitskiy2020image} & 0.9430 & 0.6436 & 0.6068 & 0.6655 & 0.7371 \\
        RETFound~\cite{zhou2023foundation} & 0.9599 & 0.6584 & 0.6377 & 0.7094 & 0.7465 \\
        VisionFM~\cite{qiu2024development} & 0.9716 & 0.6632 & 0.6546 & 0.7220 & 0.7365 \\
        \textbf{VAMPIRE} & \textbf{0.9742} & \textbf{0.7649} & \textbf{0.7205} & \textbf{0.7462} & \textbf{0.7668} \\
        \bottomrule
        \end{tabular}
    \label{tab:extension}
    \end{adjustbox}
    \end{minipage}
    \begin{minipage}[t]{0.49\textwidth}
    \centering
    \caption{Ablation study of the proposed MBD and IEM modules in our method on the OCTA-CVD dataset.}
    \begin{adjustbox}{width=\linewidth}
        \begin{tabular}{lccc}
        \toprule
        \textbf{Model} & \textbf{F1 Score} & \textbf{AUC} & \textbf{AUPR} \\
        \midrule
        Baseline & 0.5522\scriptsize{$\pm$0.0255} & {0.8061\scriptsize{$\pm$0.0192}} & {0.6009\scriptsize{$\pm$0.0341}} \\
        w/ MBD & 0.6162\scriptsize{$\pm$0.0322} & 0.8208\scriptsize{$\pm$0.0180} & 0.6301\scriptsize{$\pm$0.037} \\
        w/ IEM & 0.6101\scriptsize{$\pm$0.0319} & 0.8195\scriptsize{$\pm$0.0193} & 0.6398\scriptsize{$\pm$0.0355} \\
        \textbf{Both} & \textbf{0.6271\scriptsize{$\pm$0.0293}} & \textbf{0.8244\scriptsize{$\pm$0.0153}} & \textbf{0.6467\scriptsize{$\pm$0.0394}} \\
        \bottomrule
        \end{tabular}
    \label{tab:ablation}
    \end{adjustbox}
    \end{minipage}
\end{table*}

\noindent\textbf{Extended Experiment.} To further verify the effectiveness and generalizability of our method, we also collect a dataset from another hospital for evaluation, which contains 765 images from 379 patients. The AUC results of all the categories are shown in Table~\ref{tab:extension}. Notably, our proposed VAMPIRE model consistently outperforms other baselines across all five categories, demonstrating its remarkable ability to interpret OCTA-specific features with a particular target on vessel characters.

\subsection{Ablation Studies}
\textbf{Effectiveness of Proposed Modules.} The improvement of our VAMPIRE model can be attributed to the proposed two components, the Mamba-Based Directional (MBD) module modeling the trajectories of inner blood vessels and the Information-Enhanced Morphological (IEM) module combining vessel morphology descriptions with the image features. From Table~\ref{tab:ablation}, we can conclude that applying MBD and IEM independently can still improve the overall result by 6.4\%, 2.92\% in F1 score and 2.92\%, 3.89\% in AUPR. This proves that both designs significantly enhance the model performance, and removing either module would lead to decreased results.

\begin{figure}[t]
    \begin{minipage}{0.48\textwidth}
    \centering
        \includegraphics[width=0.72\textwidth]{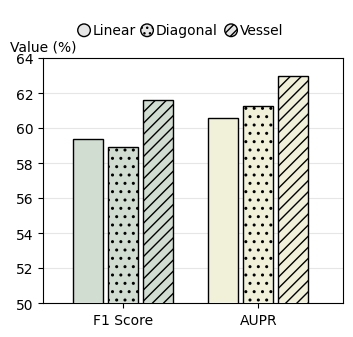}
        \caption{Comparison between Different Scanning Strategies for the Mamba-Based Directional Module.}
        \label{fig:mbd}
    \end{minipage}
    \begin{minipage}{0.48\textwidth}
    \centering
        \includegraphics[width=0.72\textwidth]{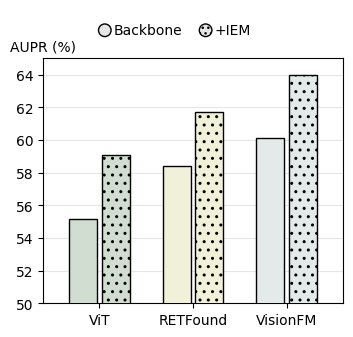}
        \caption{The Results of Integrating the Information-Enhanced Morphological Module into Different Backbones.}
        \label{fig:iem}
    \end{minipage}
\end{figure}

\noindent\textbf{Analysis of Vessel-Following Scanning Strategies.} To demonstrate the indispensability of our proposed Vessel-Following scanning strategy, we perform linear scanning and diagonal scanning for comparison. Fig.~\ref{fig:mbd} presents the average F1 score and AUPR metric of these different implementations. It can be observed that other scanning methods show minimal contribution to performance enhancement. This underscores the unique advantage of our approach in extracting specific vascular features, which particularly aligns with the OCTA imaging characteristics.

\noindent\textbf{Generalizability of Vessel Shape Information.} To prove the value of descriptive vessel shape knowledge, we similarly integrate the IEM module into other backbones and assess the impact. Fig.~\ref{fig:iem} shows that our IEM module brings performance improvements regardless of the backbone architecture employed, demonstrating the effectiveness of incorporating external knowledge about vessel morphology. This suggests that leveraging vessel shape information as prior knowledge would further enhance the OCTA-specific feature extraction process.


\section{Conclusion}
In this paper, we formulate a multi-task paradigm that simultaneously predicts CVD risk and related blood conditions, substantially enhancing risk estimation accuracy and reliability.
We introduce OCTA-CVD dataset, the first collection of OCTA enface images for joint CVD risk and factor assessment, along with a Vessel-Aware Mamba-based Prediction model with Informative Enhancement (VAMPIRE). Our framework effectively concentrates on the vital vessel structure features within the OCTA images through two key components: a Mamba-Based Directional (MBD) Module that extracts vessel trajectory features, and an Information-Enhanced Morphological (IEM) Module that incorporates vessel shape knowledge. These components collaboratively extract CVD-specific vessel features from OCTA images, significantly improving CVD risk prediction and factor estimation accuracy.

\subsubsection{Acknowledgements.} 
This work was supported by a research grant from the Joint Research Scheme (JRS) under the National Natural Science Foundation of China (NSFC) and the Research Grants Council (RGC) of Hong Kong (Project No. N\_HKUST654/24), as well as grants from the National Natural Science Foundation of China (Grant Nos. 62306254 and 82302300) and the Guangdong Science and Technology Program (Grant No. 2023A0505030004).

\subsubsection{\discintname}
The authors have no competing interests to declare that are
relevant to the content of this article.

\bibliographystyle{splncs04}
\bibliography{refs}
\end{document}